\documentclass[conference]{IEEEtran}
\IEEEoverridecommandlockouts

\usepackage{cite}
\usepackage{amsmath,amssymb,amsfonts}
\usepackage{graphicx}
\usepackage{booktabs}
\usepackage{subcaption}
\usepackage{url}
\usepackage{xurl}
\usepackage[table]{xcolor}
\usepackage{textcomp}
\usepackage{comment}
\usepackage{booktabs}
\usepackage{tabularx}
\usepackage{array}
\usepackage{enumitem}

\begin{document}
\bstctlcite{IEEEexample:BSTcontrol}
\title{Seeing at the Edge: Energy Characterization\\
of Vision-Language Model Inference\\ on Edge Platforms%
\thanks{This work was partially supported by the Interdisciplinary
Undergraduate Research Community (IURC) Grant at Hofstra University.
J. Liao and H. Kondle contributed equally to this work.}}

\author{
\IEEEauthorblockN{
Junlin Liao\textsuperscript{1,*},
Harshit Kondle\textsuperscript{2,*},
Di Liu\textsuperscript{3},
and Zonghua Gu\textsuperscript{1}
}
\IEEEauthorblockA{
\textsuperscript{1}Hofstra University, Hempstead, NY, USA\\
\textsuperscript{2}Bridgewater-Raritan High School, Bridgewater, NJ, USA\\
\textsuperscript{3}Norwegian University of Science and Technology,
Trondheim, Norway
}
}

\maketitle

\begin{abstract}
Vision-language models (VLMs) are moving onto edge devices, but little
is known about how much energy they draw on single-board computers with NPU or GPU
accelerators, or where that energy goes. We measure total-system energy,
latency, and thermal behavior of Qwen2-VL-2B-Instruct and its text-only
sibling on a Raspberry Pi~5 with a Hailo-10H NPU and on a Jetson Orin
Nano Super, on both CPU and accelerator, and separate the fixed energy of each request from the energy that grows
with output length. Because the
accelerators use different quantization and serving software, we compare
deployment stacks rather than silicon. Three findings stand out. First,
NPU offloading makes the Pi a much more efficient VLM platform: it cuts
energy per generated token by about two-thirds, leaves the host CPU
largely free, and avoids the thermal throttling that the CPU backend
reaches in sustained use. Second, an image adds both fixed and per-token
energy, but at typical response lengths generating the answer costs far
more than processing the image. Third, the Jetson GPU favors short
responses through its lower fixed cost per request, whereas the Pi NPU
favors longer ones through its lower cost per output token; their energy
curves cross near 40 output tokens.
\end{abstract}
\begin{IEEEkeywords}
edge AI, vision-language models, energy measurement, NPU, Raspberry Pi,
Hailo-10H, embedded inference
\end{IEEEkeywords}

\section{Introduction}
Multimodal foundation models~\cite{wang2024qwen2vl} are moving from the
cloud to the edge~\cite{gu2026scheduling}. Applications such as home monitoring, assistive
robotics, retail analytics, and field inspection increasingly require a
camera-equipped device to \emph{understand} what it sees, not merely
detect objects, under tight energy, thermal, and privacy budgets. A new
generation of accelerator HATs brings generative-AI-capable NPUs to
commodity single-board computers (SBCs): the Raspberry Pi AI HAT+~2
pairs the Raspberry Pi~5 with the Hailo-10H, a 40-TOPS NPU with 8~GB of
dedicated DRAM that can host LLMs and VLMs entirely
on-accelerator~\cite{rpi2026aihat2}.

Battery-powered and passively cooled devices may struggle to sustain
multi-watt CPU inference, yet published energy data for
\emph{multimodal} inference on these platforms remain scarce.
Edge-LLM-Bench~\cite{le2026edgellmbench} characterized \emph{text-only}
LLM inference on the Pi~5 + Hailo-10H and named VLM workloads as an open problem. Zhan et al.~\cite{zhan2026seeing} profiled on-device VLM energy
on GPU-class hardware and found decode-dominated energy (86--97\%). We
instead measure request-level energy at the input of complete SBC
systems. We ask how energy and latency differ across Pi-CPU, Pi-NPU,
Orin-CPU, and Orin-GPU, how much energy an image adds to a request, and
how output length changes the NPU--GPU comparison. An INA228 power
monitor~\cite{ti2022ina228,adafruit2026ina228} in series with the Pi's
5~V and the Jetson's 19~V inputs measures a paired workload,
Qwen2-VL-2B-Instruct and
Qwen2-1.5B-Instruct~\cite{wang2024qwen2vl,yang2024qwen2}, on all four
paths under a protocol that extends Edge-LLM-Bench to multimodal
prompts.

Our decode-dominance result agrees with Zhan et
al.~\cite{zhan2026seeing}, and the host-liberation benefit that
Edge-LLM-Bench~\cite{le2026edgellmbench} reported for text carries over
to multimodal workloads. The main new analysis separates fixed request
energy from energy that grows with output length and uses the two terms
to explain where the NPU and GPU stacks cross. The Hailo and llama.cpp
paths differ in quantization, compiler, runtime, and serving software,
so we compare deployed stacks (Section~\ref{sec:workload}). The CPU and GPU paths share byte-identical Q4\_K\_M GGUF weights and \texttt{llama-server}, so their comparison is the closest this study comes to a hardware-only contrast.

We make three contributions:
\begin{itemize}[
    leftmargin=1.2em,
    labelsep=0.4em,
    itemsep=1pt,
    topsep=2pt]
    \item We quantify the fixed and output-dependent energy costs of
    image-conditioned inference with an image/no-image ablation, a
    one-token proxy, a vision-token sweep, and a fixed-output-length
    experiment (Sections~\ref{sec:res-ablation}--\ref{sec:res-scaling}
    and~\ref{sec:res-crossover}).
    \item We compare four edge deployment paths under host contention
    and sustained operation, and estimate the NPU--GPU request-energy
    crossover in a controlled fixed-output probe
    (Sections~\ref{sec:res-headline}--\ref{sec:res-crossover}).
    \item We examine the quality--energy trade-off across image
    resolutions and output caps with deterministic grading
    (Section~\ref{sec:res-quality}).
\end{itemize}

\section{Related Work}
\label{sec:related}

\textbf{LLM benchmarking and energy on edge devices.}
MLPerf~\cite{reddi2020mlperf} standardized inference benchmarking before
generative workloads; since then, MELTing~point~\cite{laskaridis2024melting}
ran quantized LLMs on smartphones and Jetson-class devices,
MobileAIBench~\cite{murthy2024mobileaibench} and
PalmBench~\cite{li2025palmbench} covered task quality, latency, and
resource use on mobile platforms, Dhar et al.~\cite{dhar2024empirical}
and Lu et al.~\cite{lu2024slm} profiled edge memory/compute limits,
Samsi et al.~\cite{samsi2023words} established datacenter-scale LLM
energy methodology, and surveys~\cite{qu2025mei4llm} identify
\emph{energy transparency} as a persistent gap. MELODI~\cite{husom2024price}
finds that response length and hardware strongly predict LLM inference
energy, while prompt-complexity features have comparatively weak associations. We extend this output-length perspective to multimodal
inference on NPU-equipped SBCs, separating image-attributable fixed
energy from marginal decode energy.

Tummalapalli et al.~\cite{tummalapalli2026energy} find that thermal and
memory-bandwidth constraints, rather than peak compute, dominate
sustained 1.5B-model inference across mobile, NPU, and GPU platforms
including the Pi~5 + Hailo-10H, and Edge-LLM-Bench~\cite{le2026edgellmbench}
benchmarked text-only LLMs on the identical Pi platform; neither
measures \emph{multimodal} inference. Zhan et al.~\cite{zhan2026seeing}
gave the energy \emph{anatomy} of VLM inference on GPU-class hardware
(RTX~3070, Orin~NX; rail-level, clocks pinned): power as a model
fingerprint, decode dominance, output length as the controlling
variable, and a $\leq$10\% bound on visual-token-pruning savings. We
instead measure \emph{total-system} energy on IoT-class SBCs under DVFS
defaults, add NPU and CPU cells, and decompose by \emph{modality
condition} rather than by phase, yielding the deployed marginal cost of
image conditioning, which their visual-token bound does not address.

\textbf{Dataflow NPUs.}
\label{sec:related-npu}
Generative performance on dataflow NPUs is strongly stack-dependent and
not predicted by nominal TOPS: vendor figures favor the Hailo-10H over
the Pi CPU for Qwen2-1.5B decode~\cite{hailo2026llmwhitepaper}, whereas
independent measurements and Edge-LLM-Bench report the opposite
ordering~\cite{ailocallab2026tops,le2026edgellmbench}
(Section~\ref{sec:discussion}). These conflicts motivate measuring all
paths under a common protocol with hardware-integrated measurement.

\textbf{Efficient inference and edge VLMs.}
Systems such as T-MAC~\cite{wei2025tmac} optimize the text path,
quality--energy studies focus on quantized
LLMs~\cite{husom2025sustainable}, and edge-oriented VLMs reduce model
size and visual-token budgets~\cite{chu2023mobilevlm}; Qwen2-VL's
dynamic resolution makes image resolution an explicit deployment
variable~\cite{wang2024qwen2vl}. Existing mobile multimodal benchmarks
do not report hardware-integrated total-system energy on NPU-equipped SBCs. We measure it jointly with task quality and latency.

\section{Background and Platform}
\label{sec:background}

\subsection{Raspberry Pi 5 and AI HAT+ 2}
The Raspberry Pi~5 (BCM2712, 4$\times$Cortex-A76 @ 2.4~GHz, 8~GB
LPDDR4X) hosts the AI HAT+~2, whose Hailo-10H delivers $\approx$40~TOPS
(INT4) with 8~GB of dedicated LPDDR4X over PCIe
Gen~3~\cite{rpi2026aihat2}. Unlike the Hailo-8, it supports generative
workloads through HailoRT and its GenAI serving interfaces (a
REST-style text serving layer and a Python API for multimodal
models)~\cite{hailo2026genai}. Our harness sends the NPU both text and
image requests through an Ollama-style \texttt{/api/chat} client
interface (Section~\ref{sec:method}).

\subsection{Workload pair}
\label{sec:workload}
Qwen2-VL-2B-Instruct combines a \textbf{675M-parameter vision encoder
with a 1.5B-parameter Qwen2 language model}~\cite{wang2024qwen2vl}, and
Qwen2-1.5B-Instruct shares its decoder design (GQA, SwiGLU, tokenizer
family)~\cite{yang2024qwen2}, minimizing confounding from model family
and scale while enabling comparison between image-conditioned
inference, VLM inference without an image, and a closely matched
text-only reference; instruction tuning, multimodal positional encoding,
quantization, and compiled graphs still differ, so we read VLM--LLM differences as model-level comparisons. The vision
encoder processes the image once during prefill, after which decoding
proceeds autoregressively over the fused representation. Qwen2-VL uses
dynamic resolution, so vision-token count grows with image pixel
count~\cite{wang2024qwen2vl}. On the NPU both models run as
Hailo-compiled W4A8 HEF binaries through the Hailo GenAI serving
path; on the CPU and GPU both run as Q4\_K\_M GGUF served directly by
llama.cpp \texttt{llama-server}~\cite{gerganov2026llamacpp}. This asymmetry follows from each platform's toolchain. Throughout,
``Pi-NPU'' denotes the complete Hailo-10H + HailoRT~4.21 + W4A8 HEF
stack and ``Orin-GPU'' the Orin Nano GPU + \texttt{llama-server} CUDA
build + Q4\_K\_M GGUF stack (analogously Pi-CPU and Orin-CPU). Energy and accuracy differences are properties of these stacks.

\subsection{Inference phases}
Generative inference divides into \emph{prefill} (ingestion of the
full input context, including, for VLMs, the vision encoder pass and
multimodal projection) and \emph{decode} (sequential generation of
output tokens). Prefill is compute-parallel and, for VLMs,
vision-heavy; decode is memory-bandwidth-bound and dominates wall time
for long outputs.

\section{Measurement Setup}
\label{sec:method}

\subsection{Instrumentation}
An Adafruit INA228 breakout~\cite{adafruit2026ina228} in series with the
Pi's 5~V board input captures host, DRAM, accelerator, and peripheral
energy jointly; on the Jetson the same INA228 moves to the 19~V barrel
input and remains the source of every reported Joule. Its on-chip
accumulator integrates power in hardware at the conversion cadence,
immune to host sampling jitter, and a 50-Hz software sampler provides
cross-validation traces. Each trial records accumulated energy, mean
power, duration, mean bus voltage and current, I2C errors, and a
validity flag (positive energy delta, no I2C errors, bus voltage within
4.7--5.5~V on the Pi or 17.0--21.0~V on the Jetson, $\geq$2 accumulator
samples, and hardware/software agreement within 10\% at $\geq$80\%
sampling coverage). Idle-baseline power is recorded per resident-daemon state. Request
energy is the difference of two accumulator reads bracketing request
construction, submission, and the complete non-streaming response, so
it includes client-side API handling and localhost transport. Client wall time, measured inside the request call, is 14--23~ms shorter than this hardware window (means 21~ms on the Pi and 16~ms on the Jetson GPU stack); idle subtraction (Eq.~\ref{eq:idle}) uses the hardware window. The
Jetson's on-module INA3221 rails (\texttt{tegrastats}) are logged as
telemetry only: they exclude carrier-board, fan, and regulator losses
and under-report input power by up to $\sim$50\% before calibration
($\pm$3\% after~\cite{shalavi2023calibration}). In a side-by-side check
(10 prompts/backend), \texttt{VDD\_IN} read 0.757$\times$ (GPU) and
0.754$\times$ (CPU) of INA228 input power (CV 1.2\%/1.0\%).

\textbf{Measurement uncertainty.} We did not calibrate against a
traceable reference load. The 15~m$\Omega$ shunt has 0.1\%
tolerance~\cite{adafruit2026ina228}, and the INA228 specifies
$\pm$0.05\% gain error, $\pm$1~$\mu$V shunt offset,
$\pm$20~ppm/$^{\circ}$C gain drift, a $\pm$0.5\% oscillator setting the
accumulator time base, and 1.0\%-of-full-scale energy accuracy (all
max)~\cite{ti2022ina228}. Both platforms use the same breakout, shunt,
\texttt{SHUNT\_CAL} value, and driver configuration (averaging~16,
other settings at defaults), so the proportional terms (shunt, gain,
oscillator) cancel to first order in within-platform and Pi/Jetson
ratios but not in absolute Joules. Of the remaining terms, shunt offset
($\approx$67~$\mu$A, $<$0.01\% of the $\approx$1.3/0.7~A load
currents), gain drift ($\leq$0.02\% per 10\,$^{\circ}$C), and
bus-voltage offset ($\pm$1.0~mV typical, $\approx$0.02\% of a 5-V and
$\approx$0.005\% of a 19-V reading) are small. Because VBUS is sensed on
the supply side of the shunt~\cite{adafruit2026ina228}, shunt
dissipation is included in the measured energy ($\approx$0.4\% of
Pi-NPU versus $\approx$0.06\% of Orin-GPU input power), biasing Pi
energy upward by $\approx$0.3 points; sense-to-board wiring loss was not
measured. These estimates explain why shared errors cancel. A calibrated bound on total uncertainty would require a common-load calibration: a programmable load at 5 and 19~V over the observed currents, referenced to a calibrated meter.

\subsection{Prompt suite}
\label{sec:suite}
We adopt the Edge-LLM-Bench protocol~\cite{le2026edgellmbench}: 30
prompts---10 short factual questions (10--30 input tokens), 10 medium
instruction-following tasks (50--100), and 10 long multi-turn
interactions (150--300)---with a fixed 256-token output cap and
near-deterministic decoding (temperature 0.01). For every text prompt we
construct a \emph{paired vision-language variant}: the VLM receives an
image plus an instruction referencing it, while the LLM receives the
same task with the scene described in text, and image content matches
the description exactly (e.g., a street photo with precisely four cars
pairs with ``a street scene shows three parked cars and one moving
car''). Long prompts are scripted multi-turn conversations with fixed
assistant turns, so input token counts are deterministic. Only the final response is generated and measured.

\subsection{Execution conditions and protocol}
On each platform we run the three conditions (VLM-V, VLM-T, and LLM-T)
on the CPU and on the accelerator. \texttt{llama-server}'s
OpenAI-compatible interface drives the CPU and CUDA backends, and an
Ollama-style \texttt{/api/chat} client drives the NPU for both text and
image requests. Client-observed wall time, measured the same way for
every backend, is the primary latency metric. Backend-reported prefill
and decode fields are secondary, and generated-token counts are verified by offline tokenization. A probe records each endpoint's reported timing fields and software versions. Across forced output lengths, the decode rate
implied by the wall-time slope matches the backend-reported rate within 0.2\% (e.g., 17.46 vs.\ 17.48~tok/s on the NPU and 25.57 vs.\ 25.58~tok/s on the GPU at 672~px). For a one-token request, wall time exceeds the reported prefill by 0.12~s (NPU) and 0.08~s (GPU), about one decode step plus request handling. The Hailo path does not report whether its prefill time includes
vision encoding, so cross-stack prefill comparisons are approximate.

Measurements run in sequential blocks, one per backend$\times$condition
cell: on the Pi the three NPU blocks precede the three CPU blocks, and
on the Jetson the GPU blocks precede the CPU blocks. Within a block,
prompts follow a fixed shuffled order (the same in every block). Each prompt gets one discarded warm-up request with an unrelated prompt on
the same model and modality (a failed warm-up aborts the block),
followed by three consecutive measured trials. Each trial starts only
once the SoC is below 60\,$^{\circ}$C, and the pre-trial temperature is logged. The common threshold standardizes the initial thermal state on
both platforms. Ambient is $\approx$20\,$^{\circ}$C, with the official
active cooler on the Pi and the developer-kit fan on the Jetson, both
powered through the measured input. Hosts run headless over SSH. Host CPU utilization (from \texttt{/proc/stat} deltas over all cores,
averaged within each request) and peak memory are logged at 1~Hz. A
preflight check verifies the image SHA-256 registry and prompt token
counts and runs a one-token inference on every
backend$\times$modality.

\textbf{Session-level controls.} The CPU and GPU \texttt{llama-server}
processes start and stop with each block, so trials exclude model-load
cost and the warm-up absorbs first-request effects. The Hailo daemon
stays resident for the whole Pi campaign (one HEF at a time, selected per request by model name, which is why the NPU LLM and VLM idle states differ), so the
Pi-CPU baseline is CPU execution on the HAT-equipped system with the
daemon running. Idle baselines are measured once per daemon state,
before the first block. Every \texttt{llama-server} request sets
\texttt{cache\_prompt=false}, and a fail-closed guard rejects any
measured request whose reported prompt-evaluation token count falls
below the prompt length, the signature of a prefix-cache hit. The NPU
endpoint has no cache-disable field, so the guard is its main check. Its requests carry no
context between them. Background load is visible in the per-trial host
CPU logs.

\textbf{Statistical analysis.} Repeated trials of one prompt are not
independent, and CPU/NPU measurements are paired by prompt, so the
prompt is the unit of analysis: we average repetitions per prompt and
test differences with paired $t$-tests over the 30 prompt-level means,
Holm-corrected across 15 hypotheses (three modalities $\times$ five
metrics: decode rate, energy per token, mean power, prefill time, and
energy per inference), with 95\% CIs, $d_z$ effect sizes, and Wilcoxon
signed-rank tests as a distribution-free check. Figure error bars are
95\% CIs across prompt-level means.

\subsection{Three-condition modality ablation}
\label{sec:ablation}
We run (i)~\emph{LLM-T}, the textualized-scene prompt on Qwen2-1.5B;
(ii)~\emph{VLM-T}, the identical vision prompt on Qwen2-VL-2B without
the image; and (iii)~\emph{VLM-V}, the same prompt with the image. The
image contrast $\Delta E_{\text{visual}} = E(\text{VLM-V}) -
E(\text{VLM-T})$ (likewise $\Delta t_{\text{visual}}$) compares the same
model, prompt, and decoding settings with and without the image. This
marginal image-conditioning cost covers image loading and
preprocessing, vision encoding, projection, visual-token prefill,
request overhead, and any change in output length. VLM-T is a control for the deployed VLM: its instruction refers to an image the model
does not receive and may draw a ``no image'' reply. VLM-T response texts
were not kept, so we report input tokens, generated tokens, and
termination reasons instead (Table~\ref{tab:diagnostics}) and rely on
the one-token and fixed-length experiments for the cleaner estimates of
image cost. LLM-T serves as a reference: besides the model differences above, its prompt contains an
experimenter-written scene description, so perception has already been
done outside the measured energy.

\subsection{One-token proxy and vision-token scaling}
\label{sec:proxy}
Each prompt is also run with generation capped at one token, giving the
\emph{one-token request energy} $E_{\text{one}}$ (request handling,
image decoding, tokenization, and one decode step). This campaign
repeats the 30-prompt, three-condition, two-backend design with three
measured repetitions per cell (540 requests per platform, excluding
warm-ups) as separate runs after the full-length blocks. With full-run
energy $E_{\text{full}}$, $(E_{\text{full}}-E_{\text{one}})/(N-1)$
estimates the average energy per additional output token. Because the two runs are separate executions at different times, this is a noisy
paired contrast, and negative differences are kept and flagged. The
VLM-V$-$VLM-T one-token difference is the marginal image-conditioning
energy of a one-token request, i.e., the fixed term $\alpha_b$ of
Section~\ref{sec:res-crossover} plus one step of the decode-slope
difference. For scaling,
we sweep one image/prompt pair over long-side resolutions of
224--1120~px and, separately, attach 0/1/2 copies of the same image at fixed resolution. Vision-token counts follow the model's smart-resize
rule, cross-checked against the Hugging Face processor.

\textbf{Fixed-length probe.} A follow-up run (Aug.~9) holds one medium
prompt (M01) and its image fixed and forces output lengths of 1, 8, 16,
32, 64, 128, and 256 tokens, three consecutive trials each, on the
Pi-NPU and Orin-GPU stacks (VLM-V at 448, 672, and 1120~px; VLM-T and
LLM-T once; 21 trials per path). Length is forced with
\texttt{max\_tokens}=$N$ plus \texttt{ignore\_eos} on
\texttt{llama-server} and \texttt{num\_predict}=$N$ with an empty stop
list on the NPU endpoint. All 210 trials reached the target length
(finish reason ``length''), and none was retried or excluded. Lengths run in increasing order after one discarded 8-token warm-up with an unrelated prompt, and each trial starts below 60\,$^{\circ}$C. Energy is regressed on achieved length by ordinary least squares.

\subsection{Quality--energy protocol}
\label{sec:quality-method}
Ground truth follows from the suite's controlled content
(Section~\ref{sec:suite}): staged photos match their text descriptions,
and generated document and chart images encode the same values as the
text variants, including a planted arithmetic error (a \$3.06 tax line
whose correct 9\% value is \$3.42) whose detection is itself graded. An
answer key with tolerant alternatives grades responses deterministically
by regex and numeric extraction. Twenty-five of the 30 prompts have
objective answers (counts, colors, sign text, JSON fields, rankings,
unit prices, arithmetic, root causes) and form the accuracy denominator. The five fully generative long prompts are checked for format only. A
prompt counts as correct only if every required component matches, by
majority vote (at least two of three trials) in the three-repeat native
pass and on the single response in the sweeps. In this dataset all three repetitions of every gradable prompt received the same grade in every cell, so grading each trial separately gives identical values and no conclusion depends on the majority vote. Energy per correct answer
sums the mean request energy of each gradable prompt and divides by the
number $C$ of correct prompts:
\begin{equation}
E_{\text{correct}} = \frac{1}{C}\sum_{i=1}^{25}\bar{E}_i,
\qquad
\bar{E}_i = \frac{1}{3}\sum_{j=1}^{3} E_{ij}.
\label{eq:epc}
\end{equation}
This normalizes energy per benchmark prompt. In the single-run sweeps,
$\bar{E}_i$ is the energy of the single response. Two single-run sweeps
on the accelerated cells vary image resolution (long side
448/672/1120~px, output cap 256) and output cap (32/64/128/256 tokens,
native-resolution images; resized images use Lanczos resampling and JPEG quality 90), the deployment-side counterpart of the
output-length lever studied by Zhan et al.~\cite{zhan2026seeing}. Their
intervals resample the 25 gradable prompts (2000 draws) and so reflect
prompt sampling only.

\textbf{Interference and thermal protocols.}
Following~\cite{le2026edgellmbench}, \texttt{stress-ng}
(\texttt{matrixprod}; four workers on the Pi, six on the Jetson)
saturates the host CPUs during inference on VLM-V prompts S01, M01, and L01 (three trials each, isolated and then under stress), and both inference and stress-workload throughput are recorded. For each backend, three
30-minute sustained sessions (alternating backends with 5~min of cooling,
looping over the ten medium VLM-V prompts, starting at 41--44\,$^{\circ}$C on the Pi and 45--49\,$^{\circ}$C on the Jetson) log power, temperature, clock
frequency (\texttt{vcgencmd measure\_clock arm} on the Pi, sysfs
\texttt{cpufreq} on the Jetson), and completions at 1~Hz. Pi throttling is read from the hardware \texttt{vcgencmd get\_throttled} flag; no reliable Jetson throttling flag was available, so we report Jetson temperatures and clocks.
Table~\ref{tab:testbed} summarizes the testbed.

\begin{table}[t]
\caption{Testbed configuration}
\label{tab:testbed}
\centering
\scriptsize

\begin{tabularx}{\columnwidth}{@{}
    l
    @{\hspace{1.5pt}}
    >{\raggedright\arraybackslash}X
@{}}
\toprule
\textbf{Component} & \textbf{Configuration} \\
\midrule

SBC1 &
Raspberry Pi 5 Model B Rev 1.1, BCM2712,
4$\times$A76 @ 2.4 GHz, 8 GB \\

NPU &
AI HAT+ 2 (Hailo-10H, 40 TOPS, 8 GB, PCIe Gen 3) \\

NPU stack &
HailoRT 4.21.0 (hailort 4.21.0-1, Hailo-10H firmware 4.21.0),
Hailo GenAI serving path (Ollama-compatible \texttt{/api/chat}) \\

CPU stack &
llama.cpp release b9412 (commit 7c4e9e0)
\texttt{llama-server}; launch: 4 threads, ctx-size 4096 \\

OS &
Debian GNU/Linux 12 (bookworm), Raspberry Pi OS 64-bit \\

\midrule

SBC2 &
NVIDIA Jetson Orin Nano Developer Kit (8 GB),
Ubuntu 22.04.5 LTS (JetPack 6.2, L4T R36.4.3),
MAXN SUPER mode (nvpmodel mode 2, verified),
schedutil CPU governor, \texttt{jetson\_clocks} not applied,
unpinned GPU/EMC frequency ranges; CUDA 12.6.11,
cuDNN 9.3.0, TensorRT 10.3.0; kernel 5.15.148-tegra \\

SW stacks &
same llama.cpp release b9412: CPU-only build (4 threads) and CUDA
build (compute capability 8.7, all layers offloaded) \\

\midrule

VLM &
Qwen2-VL-2B-Instruct: W4A8 HEF (Pi NPU);
Q4\_K\_M GGUF + F16 multimodal projector
(Pi CPU and Orin CPU/GPU, \texttt{llama-server}) \\

LLM &
Qwen2-1.5B-Instruct: W4A8 HEF (Pi NPU);
Q4\_K\_M GGUF
(Pi CPU and Orin CPU/GPU, \texttt{llama-server}) \\

\bottomrule
\end{tabularx}
\end{table}

\section{Results}
\label{sec:results}
Most results come from one measurement occasion per platform (Aug.\ 8,
2026), with the Jetson's MAXN SUPER mode verified; the fixed-length
probe (Section~\ref{sec:res-crossover}) ran on Aug.\ 9 under the same
protocol and power mode. Each platform contributes 540 ablation and 540
one-token trials, idle baselines, resolution and image-count sweeps,
contention runs, and three 30-minute thermal sessions per backend. The Jetson adds the INA228-versus-\texttt{tegrastats} cross-check. All
per-request trials pass the validity gate (accumulator/software
disagreement mean 2.67\%, max 4.77\%; no I2C errors or zero-token
responses). The Pi and Jetson GGUF and mmproj files are byte-identical
(SHA-256).

\subsection{Headline energy and throughput}
\label{sec:res-headline}
\begin{table}[t]
\caption{Raspberry Pi main benchmark: mean per-condition results over
30 prompts $\times$ 3 trials (prompt-level means; 95\% CIs). Energy/token
is a mean of per-prompt ratios, not total energy over total tokens.}
\label{tab:headline}
\centering
\scriptsize
\setlength{\tabcolsep}{1.7pt}
\begin{tabular}{@{}lcccc@{}}
\toprule
Metric & CPU LLM-T & CPU VLM-V & NPU LLM-T & NPU VLM-V \\
\midrule
Mean power (W) & 8.66 ($\pm$0.10) & 9.22 ($\pm$0.05) & 5.97 ($\pm$0.04) & 6.56 ($\pm$0.03) \\
Energy/token (mJ) & 1156 ($\pm$7) & 1959 ($\pm$115) & 425 ($\pm$50) & 676 ($\pm$96) \\
Energy/inference (J) & 135.3 ($\pm$34.7) & 211.6 ($\pm$49.0) & 39.6 ($\pm$8.5) & 64.2 ($\pm$12.3) \\
Wall time (s) & 15.35 ($\pm$3.88) & 22.75 ($\pm$5.22) & 6.55 ($\pm$1.38) & 9.72 ($\pm$1.83) \\
Prefill time (s) & 0.803 ($\pm$0.182) & 2.659 ($\pm$0.164) & 1.424 ($\pm$0.019) & 2.325 ($\pm$0.031) \\
Decode rate (tok/s) & 8.07 ($\pm$0.02) & 5.92 ($\pm$0.02) & 22.33 ($\pm$0.08) & 16.31 ($\pm$0.05) \\
Host CPU (\%) & 98.6 ($\pm$0.8) & 97.9 ($\pm$1.1) & 16.8 ($\pm$0.2) & 16.8 ($\pm$0.2) \\
\bottomrule
\end{tabular}
\end{table}

NPU offloading reduces VLM energy per generated token by 65.5\%
(1959$\rightarrow$676~mJ; LLM: 63.2\%, 1156$\rightarrow$425~mJ) and mean
power by 28.9\% (9.22$\rightarrow$6.56~W), while decoding 2.75$\times$
faster and finishing 2.34$\times$ sooner end-to-end. All 15 paired
CPU-vs-NPU contrasts remained significant after Holm correction
(adjusted $p < 0.001$; minimum $|d_z| = 0.85$, $n = 30$ prompts), and
Wilcoxon signed-rank tests agree in direction and significance. Idle
floors range from 4.02~W (no \texttt{llama-server} running; Hailo
daemon resident) to 4.57~W (NPU VLM stack resident). Subtracting them (Eq.~\ref{eq:idle}) raises the offloading benefit from
65.5\% (gross) to 80.4\%.

\subsection{Per-category behavior}
For VLM-V, energy per token falls with prompt length on both backends
(short$\rightarrow$long: 2340$\rightarrow$1751~mJ on CPU, $-25$\%;
998$\rightarrow$495~mJ on NPU, $-50$\%) as fixed per-request and
per-image costs are amortized over longer generations. LLM-T shows a
mild rise on CPU ($+3$\%, 1143$\rightarrow$1173~mJ, within noise) and a
sharp fall on NPU ($-45$\%, 587$\rightarrow$323~mJ): amortization is
strongest where per-request fixed costs are largest. The VLM-vs-LLM
energy ratio contracts from 2.05$\times$ (short) to 1.49$\times$ (long)
on CPU and from 1.70$\times$ to 1.53$\times$ on NPU. The ratio is largest in the short category, where fixed request costs are least
amortized.

\subsection{Modality ablation}
\label{sec:res-ablation}
On the Pi, adding an image (VLM-T$\rightarrow$VLM-V) increased energy
per token by $574 \pm 120$~mJ on CPU (+41\%) and $185 \pm 48$~mJ on NPU
(+38\%); the LLM-T$\rightarrow$VLM-T step was $229 \pm 7$ and
$66 \pm 10$~mJ (+20\%/+15\%; mean paired differences with 95\% CIs,
$n=30$ prompts). 
\begin{figure}[t]
\centering
\includegraphics[width=1.0\linewidth]{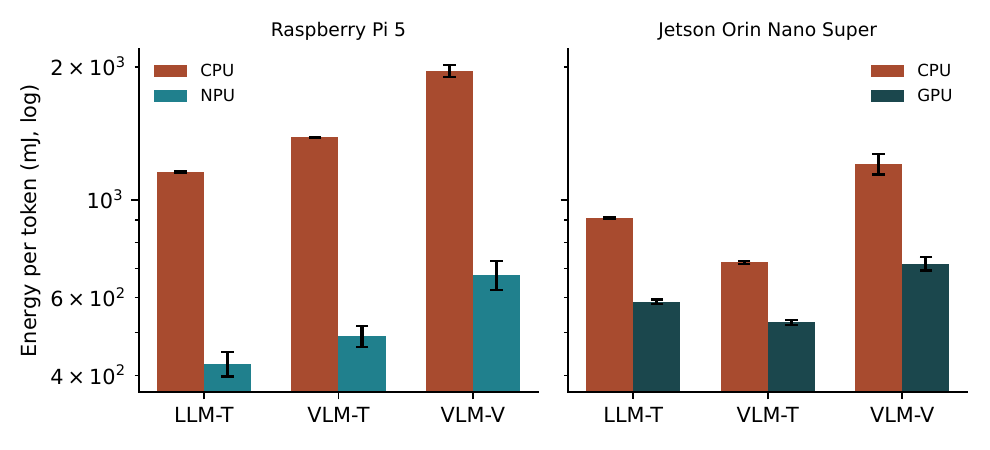}
\caption{Energy per generated token by backend and condition (prompt-level means, 95\% CI error bars, log scale). Three-condition modality ablation on the Pi (left) and Jetson (right). The LLM-T$\rightarrow$VLM-T step is positive on the Pi but negative on the Jetson, whereas image conditioning (VLM-T$\rightarrow$VLM-V) increases energy on every backend.}
\label{fig:ablation}
\end{figure}

\subsection{One-token request decomposition}
\begin{table}[t]
\caption{Per-condition request diagnostics (natural suite; prompt-level
means over 30 prompts $\times$ 3 trials). Stop reasons: end of sequence
(EOS), 256-token cap, or other.}
\label{tab:diagnostics}
\centering
\scriptsize
\setlength{\tabcolsep}{2.6pt}
\begin{tabular}{@{}llcccc@{}}
\toprule
Stack & Cond. & Text in & Visual in & Generated & EOS/cap/other (\%) \\
\midrule
Pi-CPU   & LLM-T & 125 & 0    & 117 & 97/3/0 \\
         & VLM-T & 101 & 0    & 135 & 93/7/0 \\
         & VLM-V & 101 & 1204 & 118 & 97/3/0 \\
Pi-NPU   & LLM-T & 125 & 0    & 113 & 97/3/0 \\
         & VLM-T & 101 & 0    & 128 & 93/7/0 \\
         & VLM-V & 101 & 1204 & 119 & 97/3/0 \\
Orin-CPU & LLM-T & 125 & 0    & 117 & 97/3/0 \\
         & VLM-T & 101 & 0    & 134 & 93/7/0 \\
         & VLM-V & 101 & 1204 & 120 & 97/3/0 \\
Orin-GPU & LLM-T & 125 & 0    & 116 & 97/3/0 \\
         & VLM-T & 101 & 0    & 137 & 93/7/0 \\
         & VLM-V & 101 & 1204 & 119 & 97/3/0 \\
\bottomrule
\multicolumn{6}{@{}p{0.97\columnwidth}@{}}{\footnotesize Visual-in is the API-reported prompt-token difference between VLM-V and VLM-T for the same prompt (suite mean 1204; 1120--1288 across prompts; the reported counts are identical on all four stacks). }\\
\end{tabular}
\end{table}

VLM-T did not terminate earlier than VLM-V (128--135 vs.\ 118--119
generated tokens on the Pi; Table~\ref{tab:diagnostics}), so the
one-token VLM-V$-$VLM-T contrast is the cleaner image-attributable
fixed-energy proxy. One-token request energy accounts for 11.8\% (CPU)
and 24.1\% (NPU) of per-inference VLM-V energy on average. The
marginal image-conditioning energy is 18.64~J per image on the Pi CPU
but only 6.54~J on the NPU, a 2.85$\times$ reduction (Orin: 18.42~J on
the CPU versus 7.40~J on the GPU),
smaller than the 3.29$\times$ full-run energy-per-inference ratio:
offloading benefits ingestion less than proportionately. For both
text-only conditions the NPU's one-token cost exceeds the CPU's (LLM-T
8.69 vs.\ 7.15~J; VLM-T 8.93 vs.\ 6.29~J), so the cross-backend ordering
of request overhead differs from the decode-phase ordering. The proxy
overestimates pure prefill by one decode step plus request overhead. No negative decode-energy differences occurred on either backend.

\subsection{Vision-token energy scaling}
\label{sec:res-scaling}
One-token request energy increased near-linearly with reference-processor
vision-token count across 224--1120~px (48--1200 vision tokens) on both
accelerated paths: 5.90~mJ/vision-token on the Orin GPU ($R^2=0.9974$)
and 4.82~mJ/vision-token on the Pi NPU ($R^2=0.9906$). Reducing the long
side from 1120 to 448~px reduced total one-token request energy by
$\approx$4.1$\times$ (Orin GPU) and 1.5$\times$ (Pi NPU). The smaller NPU factor reflects that stack's large non-image base cost
(Section~\ref{sec:res-crossover}), not a weaker dependence on image
size, since the two slopes are within 1.22$\times$. At every swept resolution, the API-reported prompt-token count on
both accelerated stacks equals the reference vision-token count plus a constant 34 text/template tokens (e.g., 1234 vs.\ 1200 at 1120~px). A 0/1/2-image sweep was directionally consistent: each additional image
added 2.19~J on the NPU and 2.45~J on the GPU.
\begin{figure}[t]
\centering
\includegraphics[width=0.7\linewidth]{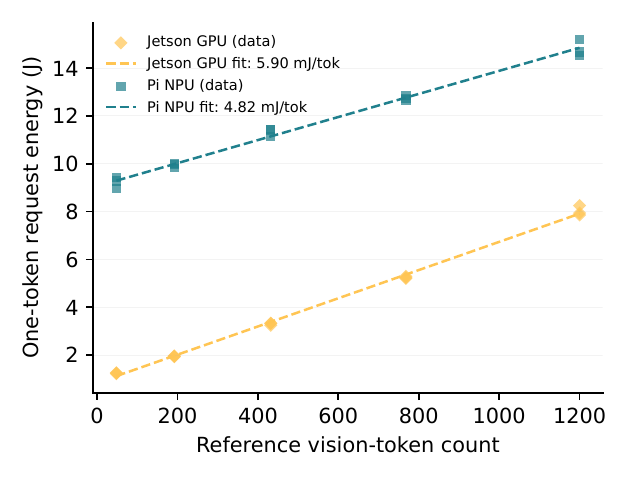}
\caption{One-token request energy vs.\ reference vision-token count (computed by the Qwen reference processor after preprocessing; level
means, 3 trials/level, linear fits). The Orin GPU and Pi NPU
slopes (5.90 and 4.82~mJ/token, respectively) are within 1.22$\times$
of each other.}
\label{fig:scaling}
\end{figure}

\subsection{Host contention}
\label{sec:res-stress}
On the Pi, saturating all four Cortex-A76 cores with \texttt{stress-ng}
cut CPU-only VLM decode throughput by 45.4\% and raised energy per token
by 110.5\%, versus 12.3\% and 37.3\% with NPU offload. \texttt{stress-ng}
in turn kept only 52.3\% of its isolated bogo-ops/s alongside CPU
inference, but 95.1\% alongside NPU inference. The contention runs did not record clock frequency, so this reduction combines scheduling and any frequency effects. The Jetson
(six workers) shows the same pattern: CPU-backend decode $-41.5$\%
(E/tok $+97.6$\%) versus GPU-backend $-8.3$\% ($+24.2$\%). Mean host CPU
utilization during inference is 87--99\% for the CPU backends versus
17\% (Pi NPU) and 3\% (Orin GPU), extending the host-liberation finding
of Edge-LLM-Bench~\cite{le2026edgellmbench} to multimodal workloads.

\subsection{Sustained thermal behavior}
\label{sec:res-thermal}
On the Jetson (three sessions per backend, MAXN SUPER verified), both
backends plateau below 65\,$^{\circ}$C (CPU
60.2--62.2\,$^{\circ}$C, peak 62.9\,$^{\circ}$C, mean power
13.5~W; GPU 62.3--64.2\,$^{\circ}$C, peak 65.0\,$^{\circ}$C, mean
power 14.4~W), with real DVFS activity (ARM clock 729.6--1510.4~MHz, GPU
clock 816--1020~MHz).

\begin{figure}[t]
\centering
\includegraphics[width=0.7\linewidth]{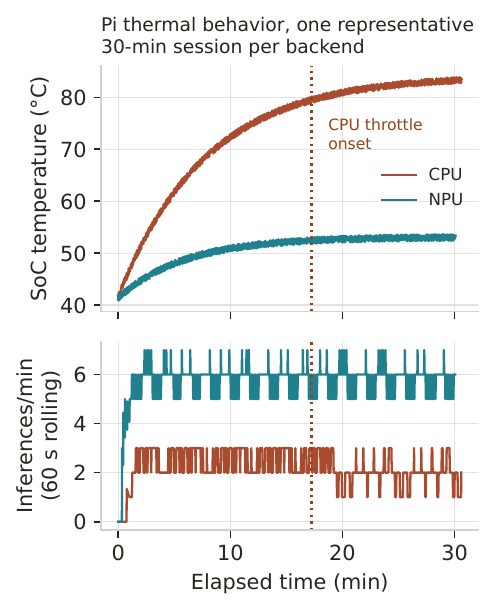}
\caption{Pi CPU vs.\ NPU temperature over one representative 30-minute
session per backend (of 3 replicated sessions each; full ranges in
Section~\ref{sec:res-thermal}), with a marker at the
CPU's throttle onset ($\approx$17~min in this session) and the aligned inference
completion rate (requests/min, 60-s rolling window) below.}
\label{fig:thermal-panel}
\end{figure}

On the Pi, the CPU backend plateaus at 83.0--84.9\,$^{\circ}$C (peak
85.9\,$^{\circ}$C), within 1\,$^{\circ}$C of the throttle
threshold, and sets the hardware throttle flag (\texttt{throttle\_status}
0x80008) in all three sessions. Throttle onset occurs 14--17~min into the sessions; afterwards the mean ARM clock falls to 1930--2007~MHz (dips to 1500~MHz, of 2400~MHz), and the recorded decode rate falls from 5.74--5.89~tok/s in the first five minutes to
4.37--4.46~tok/s in the last five ($-23.8$\% to $-24.3$\%). The NPU
backend plateaus at 53.0--55.0\,$^{\circ}$C (peak 55.6\,$^{\circ}$C,
mean power 6.7~W), stays at 2400~MHz, never sets a throttle flag, and its decode rate is
essentially unchanged over the session ($-0.2$\% to $+0.3$\%). Under identical ambient conditions, the CPU backend thus runs
at its thermal ceiling while the NPU backend keeps $>$30\,$^{\circ}$C
of headroom (Fig.~\ref{fig:thermal-panel}).

\subsection{Cross-platform comparison: Jetson Orin Nano Super}
\label{sec:res-crossplatform}
\begin{table}[t]
\caption{Cross-platform comparison (VLM-V; prompt-level means, campaigns
matched by prompt; energy/token is a mean of per-prompt ratios as in
Table~\ref{tab:headline}). Bold marks the better value where the
direction is unambiguous; $\dagger$: difference not significant
(Section~\ref{sec:res-crossplatform}).}
\label{tab:crossplatform}
\centering
\scriptsize
\setlength{\tabcolsep}{3.5pt}
\begin{tabular}{@{}lcccc@{}}
\toprule
 & Pi CPU & Pi NPU & Orin CPU & Orin GPU \\
\midrule
Mean power (W) & 9.22 & \textbf{6.56} & 12.97 & 13.50 \\
Energy/token (mJ) & 1959 & 676$^\dagger$ & 1206 & 717$^\dagger$ \\
Energy/inference (J) & 211.6 & \textbf{64.2} & 123.8 & 77.6 \\
Decode (tok/s) & 5.92 & 16.31 & 16.05 & \textbf{24.88} \\
Wall time (s) & 22.75 & 9.72 & 9.41 & \textbf{5.58} \\
Prefill (s) & 2.659 & 2.325 & 1.882 & \textbf{0.741} \\
1-token image cost, marginal (J) & 18.64 & \textbf{6.54} & 18.42 & 7.40 \\
LLM-T energy/token (mJ) & 1156 & \textbf{425} & 909 & 588 \\
Host CPU (\%) & 97.9 & 16.8 & 97.0 & 3.0 \\
Idle floor, baseline daemon state (W) & \multicolumn{2}{c}{4.02} & \multicolumn{2}{c}{5.70} \\
\bottomrule
\end{tabular}
\end{table}
\begin{figure}[t]
\centering
\includegraphics[width=0.7\linewidth]{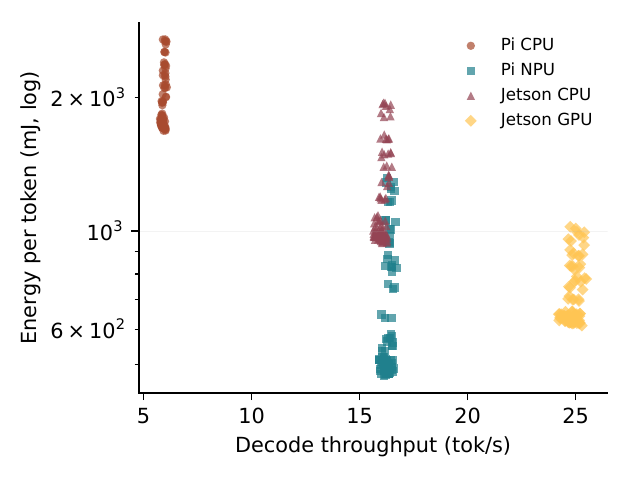}
\caption{Cross-platform throughput--energy trade-off for image-conditioned
VLM inference (VLM-V). Each point is one prompt's mean over three trials (30 points per
stack). The Pi-NPU and Orin-GPU stacks have similar mean per-prompt energy per
token, and the Orin-GPU stack decodes faster. Which stack uses less energy
depends on output length (Section~\ref{sec:res-crossover}). Logarithmic
energy axis.}
\label{fig:crossplatform}
\end{figure}

Table~\ref{tab:crossplatform} compares the four deployment stacks under
the matched protocol. Decode throughput follows Orin GPU \(>\) Pi NPU
\(\approx\) Orin CPU \(\gg\) Pi CPU (24.9/16.3/16.1/5.9~tok/s). The Orin-GPU stack decodes 1.53\(\times\) faster than the Pi-NPU stack and
completes requests 1.74\(\times\) sooner. Mean per-prompt energy per generated token was similar for the two
accelerator stacks (676 vs.\ 717~mJ): the prompt-paired difference is
$-40$~mJ (95\% CI $[-92, +11]$~mJ, $p=0.12$), so neither stack has a
demonstrably lower mean. The comparison does depend on response length:
per-token energy favors the Orin-GPU stack by 1.15$\times$ on short
prompts but the Pi-NPU stack by 1.21$\times$ and 1.28$\times$ on medium
and long prompts.

\subsection{Image-ingestion/decode and platform crossovers}
\label{sec:res-crossover}
To separate fixed request cost from marginal decode cost, we hold one
prompt and image fixed at each of three resolutions, force output
lengths of $1, 8, 16, 32, 64, 128$, and $256$ tokens (21 hardware-valid
trials per fitted path), and fit $E_b(R,N) = A_b(R) + \beta_b(R) N$ for
backend $b$, resolution $R$, and achieved output length $N$, where
$A_b(R)$ is the \emph{complete} VLM-image request intercept and
$\beta_b(R)$ the marginal decode energy per output token. With the
matching VLM-T fit, the intercept splits into a non-image base and an
image-attributable term:
\begin{equation}
\resizebox{0.98\linewidth}{!}{$
\underbrace{E_b(R,N)}_{\text{request energy}} =
\underbrace{A_{b,\text{VLM-T}}}_{\text{base (non-image)}} +
\underbrace{\alpha_b(R)}_{\text{image-attributable}} +
\underbrace{\beta_b(R)N}_{\text{decode}}
$}
\label{eq:decomposition}
\end{equation}
with $A_b(R) = A_{b,\text{VLM-T}} + \alpha_b(R)$
(Table~\ref{tab:fixed-length}; $R^2\geq0.9997$ for all eight VLM-V and
VLM-T fits, no output-length mismatches). Here the base term is the fitted non-image request intercept, including fixed prompt-processing and serving costs,
$\alpha$ the fixed energy attributable to adding an image at
resolution $R$, and $\beta$ the marginal energy per output token (the
VLM-V slope).

\textbf{Image conditioning raises the decode slope.} The VLM-V slope
exceeds the VLM-T slope ($\beta_{b,\text{VLM-T}}=361$~mJ/token on the
Pi NPU and 549~mJ/token on the Orin GPU) by $\Delta\beta_b(R)$ of 22.6 [18.8, 26.3], 24.4 [21.1, 28.6], and 42.9 [40.9, 44.7]~mJ/token on the Pi NPU and 6.9 [3.0, 10.6], 14.1 [9.1, 19.6], and 33.6 [27.7, 40.2]~mJ/token on the Orin GPU at 448, 672, and 1120~px (bootstrap 95\% CIs),
consistent with decoding over a longer multimodal context. The
image/no-image energy difference at output length $N$ is therefore
$\alpha_b(R)+\Delta\beta_b(R)N$: $\alpha$ is its fixed part, and the
slope increment counts toward both the image-attributable and the decode energy. At $N=120$ the difference is 3.5, 4.7, and 10.1~J on the Pi NPU and 1.8, 3.8, and 9.9~J on the Orin GPU (448/672/1120~px), of which the slope term contributes 51--78\% and 41--45\%, respectively.

Across image long sides of 448--1120~px, the Orin-GPU stack's fixed
request intercept is 51--81\% lower than the Pi-NPU stack's
(1.71--6.59~J versus 9.23--13.41~J), whereas the Pi-NPU stack's marginal
decode cost is 30--31.5\% lower (384--404 versus 556--583~mJ per output
token; $\beta_{\text{NPU}}/\beta_{\text{GPU}} = 0.685$--$0.694$, with
non-overlapping bootstrap 95\% CIs at every resolution). The intercept
gap does not come from image ingestion: the Orin-GPU stack's $\alpha$ is
slightly \emph{higher} at every resolution ($-$0.3 to $-$1.0~J
difference), and its advantage lies in the non-image base term (0.70
vs.\ 8.49~J, a 7.8~J gap at every resolution).

\begin{figure}[htb]
\centering
\includegraphics[width=\linewidth]{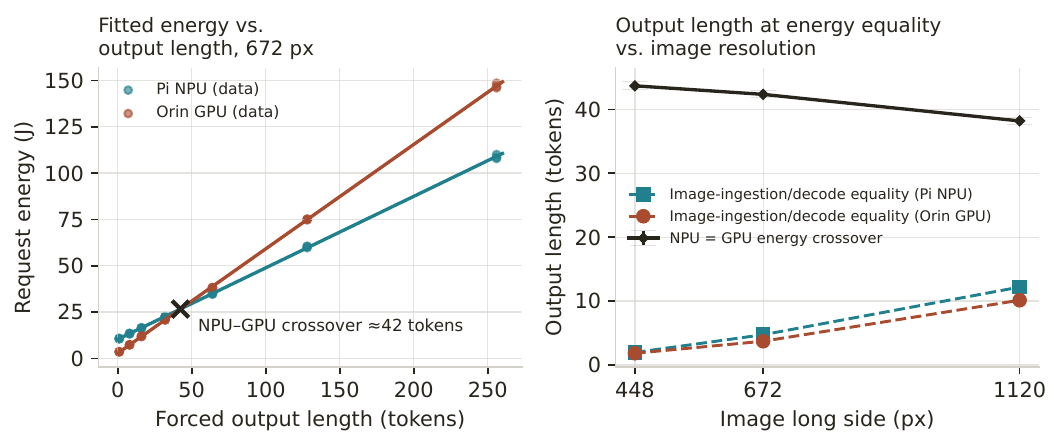}
\caption{\textbf{Left:} fitted total request energy versus forced output
length at 672~px, with the cross-platform NPU=GPU energy crossover
marked. \textbf{Right:} two distinct output-length equalities versus
image resolution. Squares and circles show the within-backend length
at which image-attributable fixed energy equals autoregressive decode energy. Diamonds show the cross-platform length at which total NPU
and GPU request energy are equal. Error bars denote the
bootstrap 95\% CI for the cross-platform crossover.}
\label{fig:crossover-panel}
\end{figure}

\textbf{Controlled-probe crossover.} Equating the fitted lines gives
$N_{\text{NPU=GPU}}(R) = (A_{\text{NPU}}(R) - A_{\text{GPU}}(R)) /
(\beta_{\text{GPU}}(R) - \beta_{\text{NPU}}(R))$: about 44, 42, and 38
output tokens at 448, 672, and 1120~px (bootstrap 95\% CIs in
Table~\ref{tab:fixed-length}; stratified by target length, 2000 draws).
This crossover comes from one prompt/image pair per resolution. As a
consistency check on the natural suite, we fit a two-point energy line per prompt $i$ and stack $b$ from the mean one-token and full-run VLM-V energies, $\hat{\beta}_{ib} = (\bar{E}^{\text{full}}_{ib} - \bar{E}^{\text{one}}_{ib})/(\bar{N}^{\text{full}}_{ib} - 1)$ and $\hat{A}_{ib} = \bar{E}^{\text{one}}_{ib} - \hat{\beta}_{ib}$, and solve $\hat{N}_i = (\hat{A}_{i,\text{NPU}} - \hat{A}_{i,\text{GPU}})/(\hat{\beta}_{i,\text{GPU}} - \hat{\beta}_{i,\text{NPU}})$. All 30 prompts give positive crossovers ($\hat{\beta}_{\text{GPU}} > \hat{\beta}_{\text{NPU}}$ for every prompt),
with a median of 40 tokens (IQR 36--46, range 33--55), around the
probe's 38--44 tokens. Below the crossover the Orin-GPU stack uses less
energy because of its lower fixed cost; above it the Pi-NPU stack uses
less because of its lower marginal cost. This matches the natural-suite
reversal (Section~\ref{sec:res-crossplatform}): short-category responses
average 27.8/27.7 generated tokens on the Pi-NPU/Orin-GPU stacks, below
the crossover, while medium (130.1/128.5) and long (200.3/201.0)
responses lie well above it. Fig.~\ref{fig:crossover-panel} plots the
fitted lines (left) and both crossover quantities (right).

\textbf{Image-ingestion/decode equality.} Within one stack, the fixed
image energy equals the decode energy of the image-conditioned request
at $N^{*}_{b}(R)=\alpha_b(R)/\beta_b(R)$: 1.94, 4.72, and 12.18 tokens
on the Pi NPU and 1.82, 3.70, and 10.12 on the Orin GPU at 448, 672, and
1120~px. Including the slope increment, at 120 tokens, the typical
response length in our suite (Table~\ref{tab:diagnostics}), the no-image
decode energy (43.4~J on the Pi NPU, 65.9~J on the Orin GPU) is
4.3--12.5$\times$ and 6.6--35.8$\times$ the image/no-image difference,
so decode energy dominates on both accelerator stacks, consistent with
Zhan et al.'s ``seeing'' versus ``speaking'' framing~\cite{zhan2026seeing}.
The two crossover quantities answer different questions:
$N^*_b$ compares fixed image energy with decode energy within one stack
and rises from about 2 to 12 tokens with resolution, whereas
$N_{\text{NPU=GPU}}$ compares total request energy across stacks and
stays near 38--44 tokens (Fig.~\ref{fig:crossover-panel}, right).
Across the factorial sweep, mean input power varied by less than 5\%
within each accelerated path (CV 1.17--3.06\%), so request duration
drives within-path energy variation.

\begin{figure}[t]
\centering
\includegraphics[width=\linewidth]{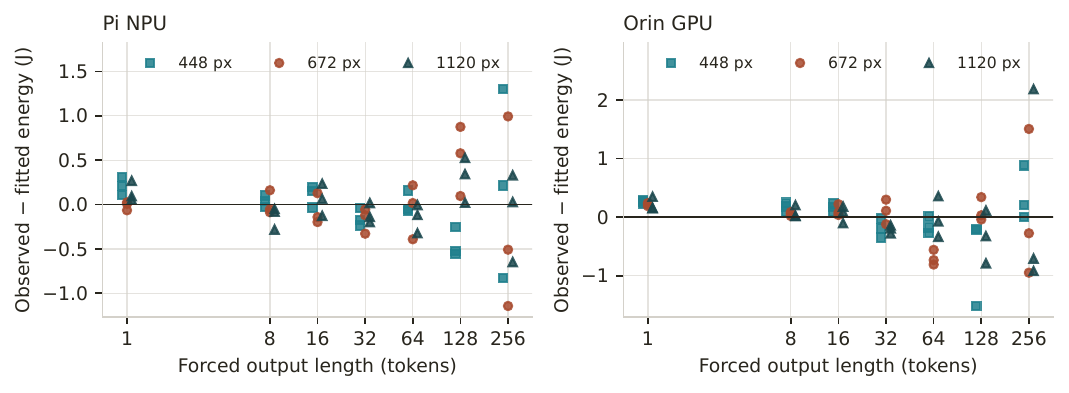}
\caption{Observed minus fitted request energy for the fixed-length
VLM-V fits (ordinary least squares on achieved length; 21 trials per
fit) on the Pi NPU (left) and Orin GPU (right) at three image
resolutions; markers are offset horizontally within each length for
legibility.}
\label{fig:residuals}
\end{figure}

\textbf{Fit diagnostics.} Fig.~\ref{fig:residuals} plots observed minus
fitted energy for the six VLM-V fits. On the Pi NPU, mean residuals at
each target length are at most 0.52~J and 2.2\% of the fitted energy. On
the Orin GPU they are at most 0.70~J, but largest in relative terms at
short lengths (up to 11.7\% at one token and 2.9\% at eight tokens) and
positive at one token in all three fits (+0.22 to +0.26~J): the line
slightly under-predicts single-token requests. At 5.6--5.8 tokens per
joule of intercept gap, an intercept bias of this size would shift the
crossover by roughly 1.4 tokens, more than the bootstrap half-widths
(about 0.6 tokens), which therefore reflect trial-resampling
uncertainty only.
\begin{table*}[t]
\centering
\caption{Fixed-length image-ingestion/decode and platform-crossover fits.
$A$ is the complete VLM-image request intercept, $\beta$ is marginal
decode energy (95\% CI, bootstrap, in brackets), $\alpha$ is the fixed
image-attributable energy ($\alpha=A_{\text{VLM-V}}-A_{\text{VLM-T}}$),
$N^*_{\text{img/dec}} = \alpha/\beta$ is the output length where
image-attributable fixed energy equals marginal decode energy, and
$N^*_{\text{NPU=GPU}}$ is the cross-platform energy crossover (95\% CI,
trial-resampling bootstrap). All fits $R^2\geq0.9997$, $n=21$
trials/path.}
\label{tab:fixed-length}
\small
\setlength{\tabcolsep}{4pt}
\begin{tabular}{@{}r rr rr r r@{}}
\toprule
& \multicolumn{2}{c}{Pi NPU} & \multicolumn{2}{c}{Orin GPU} & & \\
\cmidrule(lr){2-3}\cmidrule(lr){4-5}
Res.\ (px) & $A$ (J), $\alpha$ (J) & $\beta$ [95\% CI] (mJ/tok) & $A$ (J), $\alpha$ (J) & $\beta$ [95\% CI] (mJ/tok) &
$N^*_{\text{img/dec}}$ (NPU/GPU, tok) & $N^*_{\text{NPU=GPU}}$ [95\% CI] (tok) \\
\midrule
448  & 9.23,\ 0.74  & 384 [380.2, 387.7]  & 1.71, 1.01 & 556 [554.5, 557.8] & 1.94 / 1.82  & 43.7 [43.1, 44.3] \\
672  & 10.31, 1.82 & 386 [382.4, 389.8] & 2.78, 2.09 & 563 [559.6, 568.2] & 4.72 / 3.70  & 42.3 [41.8, 42.9] \\
1120 & 13.41, 4.92 & 404 [402.3, 405.8] & 6.59, 5.90 & 583 [578.6, 589.9] & 12.18 / 10.12 & 38.2 [37.6, 38.7] \\
\bottomrule
\end{tabular}
\end{table*}

\subsection{Exploratory quality--energy evaluation}
\label{sec:res-quality}
We compare accuracy and energy on the 25 gradable prompts in a
three-repeat native pass on all four stacks and in single-run
resolution and output-cap sweeps on the accelerators. Each sweep setting
has one run per prompt, and accuracy moves in 4-point (one-prompt)
steps, so the sweeps are exploratory.

In the native pass, the accelerator backends used less VLM energy per
correct answer than their CPU counterparts: 92.1 versus 288.7~J on the
Pi and 101.2 versus 173.1~J on the Jetson. VLM accuracy was 60\% on the
Pi CPU, Pi NPU, and Orin CPU and 64\% on the Orin GPU; LLM-T accuracy was 76\% on the first three paths and 80\% on the Orin GPU. Each 4-point difference is a single prompt. LLM-T works from an experimenter-written scene description rather than the image.

\begin{figure}[t]
\centering
\includegraphics[width=\linewidth]{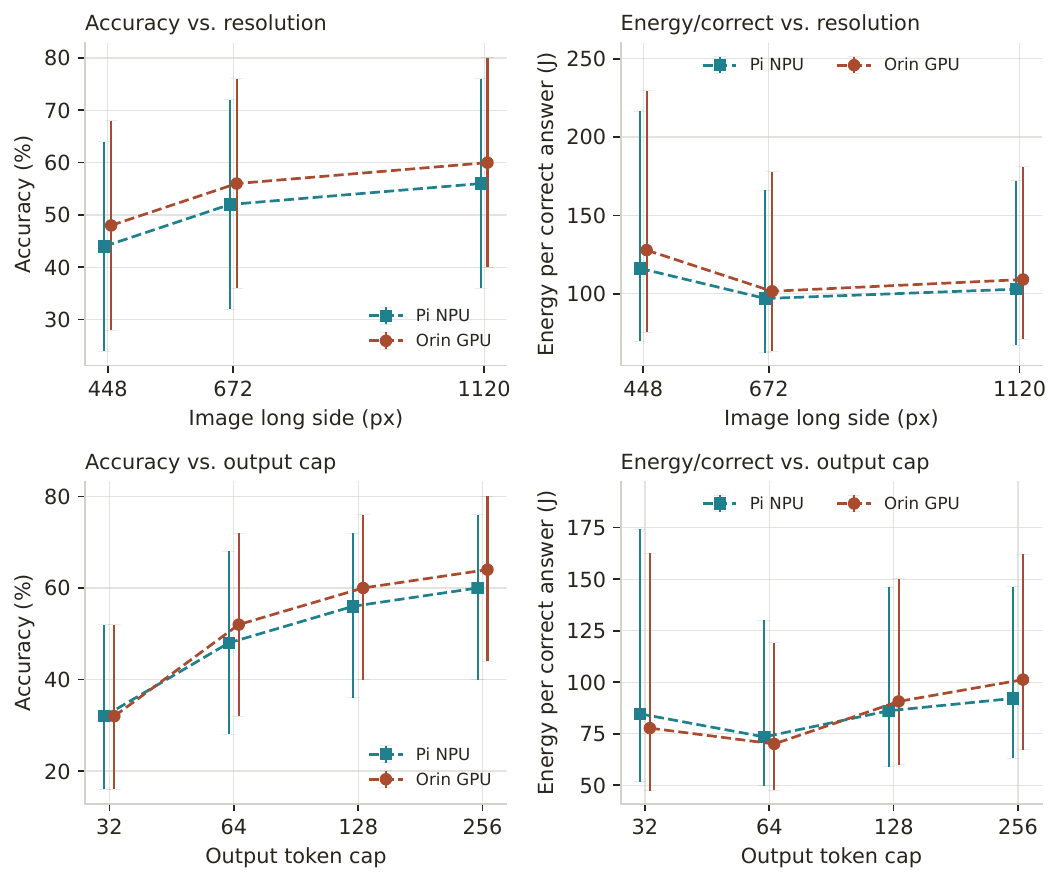}
\caption{Exploratory quality--energy sweeps: accuracy and energy per correct answer
vs.\ resolution (top) and output cap (bottom), Pi NPU and Orin GPU.
Single run per setting on 25 gradable prompts; error bars are
prompt-level bootstrap 95\% intervals (markers offset horizontally for
legibility).}
\label{fig:quality-control}
\end{figure}

In the single-run sweeps (Fig.~\ref{fig:quality-control}), accuracy rose
with both image resolution and output cap, whereas energy per correct
answer was non-monotonic. \textbf{Resolution:} at 448/672/1120~px,
accuracy was 48/56/60\% (Jetson) and 44/52/56\% (Pi), mean energy per
inference 61.4/56.9/65.5~J and 51.1/50.4/57.6~J, and energy per correct
answer 127.9/101.6/109.1~J and 116.2/97.0/102.9~J, lowest at 672~px on
both platforms. That minimum is driven mainly by the number of correct
prompts: on the Pi, mean energy per inference differs by only 1.4\%
between 448 and 672~px, while 13 versus 11 prompts were correct
(Jetson: 14 versus 12). \textbf{Output cap:} at 32/64/128/256 tokens,
accuracy was 32/52/60/64\% (Jetson) and 32/48/56/60\% (Pi), and energy
per correct answer 77.8/70.1/90.7/101.3~J and 84.7/73.5/86.2/92.2~J,
lowest at 64 tokens on both. Implied mean energy per inference (energy
per correct answer $\times$ accuracy) rises monotonically with the cap
(Pi: 27/35/48/55~J), while correct prompts rise from 8 to 12 (Pi) and
from 8 to 13 (Jetson) between the 32- and 64-token caps, consistent with
truncation of answers whose typical length is about 120 tokens (Table~\ref{tab:diagnostics}).

Prompt-level bootstrap 95\% intervals (Table~\ref{tab:sweep-ci}; error bars in Fig.~\ref{fig:quality-control}) overlap between adjacent settings; at
672~px, for example, Pi-NPU accuracy is 52\% [32, 72] and energy per
correct answer 97.0~J [62, 166]. The lowest observed values thus occur at 672~px and a 64-token cap. The 256-token Pi-NPU sweep point (92.2~J per correct answer) is a new run, separate from the native pass (92.1~J).
\begin{table}[t]
\caption{Exploratory sweeps with prompt-level bootstrap 95\% intervals
(25 gradable prompts resampled with replacement, 2000 draws; single run
per prompt and setting). Accuracy in \%; energy per correct answer in J.
The intervals capture prompt sampling only.}
\label{tab:sweep-ci}
\centering
\footnotesize
\setlength{\tabcolsep}{2.6pt}
\begin{tabular}{@{}lcccc@{}}
\toprule
 & \multicolumn{2}{c}{Pi NPU} & \multicolumn{2}{c}{Orin GPU} \\
\cmidrule(lr){2-3}\cmidrule(lr){4-5}
Setting & Accuracy & J/correct & Accuracy & J/correct \\
\midrule
448~px & 44 [24, 64] & 116.2 [70, 216] & 48 [28, 68] & 127.9 [76, 229] \\
672~px & 52 [32, 72] & 97.0 [62, 166] & 56 [36, 76] & 101.6 [63, 178] \\
1120~px & 56 [36, 76] & 102.9 [67, 172] & 60 [40, 80] & 109.1 [71, 181] \\
\midrule
32~tok & 32 [16, 52] & 84.7 [52, 174] & 32 [16, 52] & 77.8 [47, 162] \\
64~tok & 48 [28, 68] & 73.5 [50, 130] & 52 [32, 72] & 70.1 [47, 119] \\
128~tok & 56 [36, 72] & 86.2 [59, 146] & 60 [40, 76] & 90.7 [60, 150] \\
256~tok & 60 [40, 76] & 92.2 [63, 146] & 64 [44, 80] & 101.3 [67, 162] \\
\bottomrule
\end{tabular}
\end{table}

\section{Discussion}
\label{sec:discussion}

\textbf{Comparison with the literature.}
Le and Durmaz Incel~\cite{le2026edgellmbench} measured CPU decode
\emph{faster} than NPU (12.7 vs.\ 7.8~tok/s for Qwen2-1.5B); our
campaign reverses this ordering (8.1 vs.\ 22.3~tok/s, 2.77$\times$
paired). Third-party measurements report the CPU-faster ordering (9.72
vs.\ 7.05~tok/s for Qwen2.5-1.5B, decode only)~\cite{ailocallab2026tops}
and vendor figures the NPU-faster result (9.45~tok/s for
Qwen2-1.5B-Instruct)~\cite{hailo2026llmwhitepaper}.

\textbf{Idle-floor accounting.} Gross energy $E_{\text{gross}}$, our
primary quantity, is what an always-on device draws from its supply. Idle-subtracted \emph{active} energy is
\begin{equation}
E_{\text{active}} = E_{\text{gross}} - P_{\text{idle}} \cdot t,
\label{eq:idle}
\end{equation}
with $P_{\text{idle}}$ the matching resident-daemon idle power
(Section~\ref{sec:res-headline}) and $t$ the trial duration. For VLM-V
this gives $E_{\text{active}}/\text{tok} = 1035$~mJ on the CPU (gross
1959) and 202~mJ on the NPU (gross 676): the NPU-vs-CPU reduction is
65.5\% gross but 80.4\% active, because idle power makes up a larger
share of the NPU's gross energy ($\approx$70\% vs.\ $\approx$47\%). Both figures are valid, so energy reports should state which one they use.

\textbf{Implications for deployment.}
(1)~Vision-token scaling is near-linear on both accelerator stacks
($\approx$5--6~mJ/token; CPU backends not swept), so resolution is a
controllable ingestion-energy factor, but energy per
\emph{correct answer} had its lowest \emph{observed} value at an
interior resolution (Section~\ref{sec:res-quality}), so minimizing
ingestion energy alone can move away from it. (2)~The image-conditioning increment (+38--41\% on the Pi, +36--67\% on the Jetson) weighs most on short image queries (2.05$\times$ the text sibling on the Pi CPU). (3)~Accelerator-stack choice depends on response length: mean
per-token energy was similar, the Pi-NPU stack drew about half the input
power (6.6~W vs.\ 13.0--13.5~W for the Jetson stacks), and the Orin-GPU
stack decoded 1.53$\times$ faster and used less energy below the
38--44-token probe crossover. Offloading leaves $\approx$83\% of the
Pi's host CPU free (NPU) and $\approx$97\% of the Jetson's (GPU), versus
$\approx$1--13\% under the CPU backends.

\textbf{Limitations.} The results cover one VLM family at one scale
(2B), one unit of each platform, and mostly one measurement occasion
(the fixed-length probe was a separate follow-up run). Unit-to-unit
variation in SoC leakage, DRAM, and supply efficiency, other model
families and sizes, and other toolchain versions could shift the fitted
constants ($\alpha$, $\beta$, $A$, crossover lengths, vision-token
slopes), which describe these stacks on this occasion. Conditions ran in
sequential blocks, so slow drifts are not randomized across them, and
our intervals cover prompt-level variation only. Quantization and
serving differences prevent attributing energy or accuracy differences
to silicon. The image cost includes more than the vision encoder; the Hailo path accepts only complete image$+$text requests, so the encoder cannot be run on its own there. Grading is automatic (regex and numeric extraction) and was not checked by human raters on every response, so unconventional but correct answers may be misclassified. Absolute Joules are uncalibrated,
and the crossover rests on one prompt/image probe plus the natural-suite
consistency check.

\section{Conclusion}
\label{sec:conclusion}
We characterized total-system energy, latency, and thermal behavior of
VLM inference on a Raspberry Pi~5 with a Hailo-10H NPU and on a Jetson
Orin Nano Super. Three findings stand out:
\begin{itemize}[
    leftmargin=1.2em,
    labelsep=0.4em,
    itemsep=1pt,
    topsep=2pt]
    \item \textbf{NPU offloading improves Pi deployment efficiency.}
    Relative to the Pi CPU backend, the Pi-NPU stack reduces VLM energy
    per generated token by 65.5\% (1959$\rightarrow$676~mJ) and mean
    power by 28.9\% and raises decode throughput 2.75$\times$; across
    three 30-minute sessions it never sets the throttle flag that the
    CPU backend sets in every session, and it runs
    $\approx$30\,$^{\circ}$C cooler.
    \item \textbf{An image adds fixed and per-token energy, but decode
    dominates at typical response lengths.} Adding an image raises
    energy per token by 36--67\%, and one-token energy grows
    near-linearly with vision-token count (4.82--5.90~mJ/token;
    \(R^2\geq0.99\)). In the fixed-length probe, decode energy exceeds
    the fixed image energy beyond $\approx$2--12 output tokens. At the typical $\approx$120 tokens, the no-image decode term is
    4--36$\times$ the whole image/no-image increment.
    \item \textbf{The NPU and GPU stacks favor different response lengths.} Mean per-token energy was similar (676 vs.\
    717~mJ), while the Orin-GPU stack decoded 1.53$\times$ faster. The
    Orin-GPU stack's 51--81\% lower fixed request intercept and the
    Pi-NPU stack's 30--31.5\% lower marginal decode cost explain the
    length-dependent reversal, with the probe crossover near 38--44
    output tokens.
\end{itemize}
For edge VLMs, the better accelerator stack therefore depends on response
length, while on the Pi, NPU offloading improves energy, throughput,
contention resilience, and thermal headroom together. An exploratory single-run sweep found the lowest energy per correct
answer at an intermediate image resolution (672~px) and output cap
(64 tokens). Natural next steps are encoder-level energy on stacks
that expose the vision encoder, other model families and sizes, and
repeated quality sweeps to test the 672-px and 64-token minima.

\IEEEtriggeratref{3}
\bibliographystyle{IEEEtran}
\bibliography{references}

\end{document}